\documentclass[sigconf]{acmart}
\AtBeginDocument{%
  }
\usepackage{xcolor}
\usepackage[table]{xcolor}
\usepackage{pifont}
\usepackage{tabularx}
\usepackage{booktabs}
\usepackage{float}
\usepackage{hyperref}

\definecolor{topblue}{RGB}{186,85,211}
\definecolor{qcnetpurple}{RGB}{176, 118, 168}
\definecolor{demopurple}{RGB}{122, 132, 196}
\newcommand{\highlight}[1]{\textcolor{topblue}{#1}}
\setcopyright{acmlicensed}
\copyrightyear{2026}
\acmYear{2026}
\setcopyright{cc}
\setcctype{by}
\acmConference[MM '26]{Proceedings of the 34th ACM International Conference on Multimedia}{November 10--14, 2026}{Rio de Janeiro, Brazil}
\acmBooktitle{Proceedings of the 34th ACM International Conference on Multimedia (MM '26), November 10--14, 2026, Rio de Janeiro, Brazil}
\acmDOI{10.1145/3767308.3836412}
\acmISBN{979-8-4007-2213-4/2026/11}

\begin{document}

\title{A Unified Framework for Trajectory Prediction with Explicit Planning and Reaction Decomposition}

\author{Jiaheng Chen}
\affiliation{%
  \institution{Software College, Northeastern University}
  \city{Shenyang}
  \country{China}}
\email{20236778@stu.neu.edu.cn}

\author{Jiaxing Li}
\affiliation{%
  \institution{Software College, Northeastern University}
  \city{Shenyang}
  \country{China}}
\email{20237096@stu.neu.edu.cn}

\author{Tinghe Zhang}
\affiliation{%
  \institution{Software College, Northeastern University}
  \city{Shenyang}
  \country{China}}
\email{zhangtinghe5@gmail.com}

\author{Chaopeng Guo}
\correspondingauthor
\affiliation{%
  \institution{Software College, Northeastern University}
  \city{Shenyang}
  \country{China}}
\email{guochaopeng@swc.neu.edu.cn}


\begin{abstract}
  Trajectory prediction has shifted toward structured formulations with explicit social modeling. However, existing methods inadequately distinguish the functional roles of social influence in trajectory planning. Observing that agents typically form motion plans by anticipating others' future behaviors before making local reactive adjustments, we identify social interactions as playing staged roles, namely planning precedes reaction. We propose INTraJ, a unified framework that decomposes social influence into two stages: a planning stage constructs reference trajectories using future social information, and a reaction stage recovers local adjustments from the residual between full-context prediction and the reference. INTraJ supports both multi-target and single-target paradigms. Extensive experiments on four standard benchmarks, including Argoverse 2, Argoverse 2-ped, ETH/UCY, and SDD, demonstrate consistent improvements, particularly in FDE and long-horizon consistency, with state-of-the-art performance achieved in several settings. INTraJ reframes trajectory prediction as a planning-driven two-stage process, validating that staged social modeling is critical for stable predictions. The code is publicly available at \url{https://github.com/11isnotavailable/INTraJ}.
\end{abstract}




\maketitle

\section{Introduction}



Trajectory prediction estimates the future motions of agents in a scene. It supports interaction understanding and decision-making in applications such as autonomous driving and crowd analysis.~\cite{rudenko_2020_human} Although recent methods have substantially improved predictive capacity, accurate modeling remains difficult in complex multi-agent environments. 
Most mainstream approaches still adopt a coupled formulation \cite{zhou_2023_querycentric,shi_2022_motion}, encoding individual motion tendencies, scene constraints, and social interactions into a shared latent space for end-to-end prediction.
While this design provides dense information flow and strong expressiveness, it entangles social effects from different sources and time scales, increasing complexity and vulnerability to spuriously correlated neighbor patterns~\cite{pourkeshavarz_2024_cadet}.

\begin{figure}[t]
  \centering
  \includegraphics[width=\linewidth]{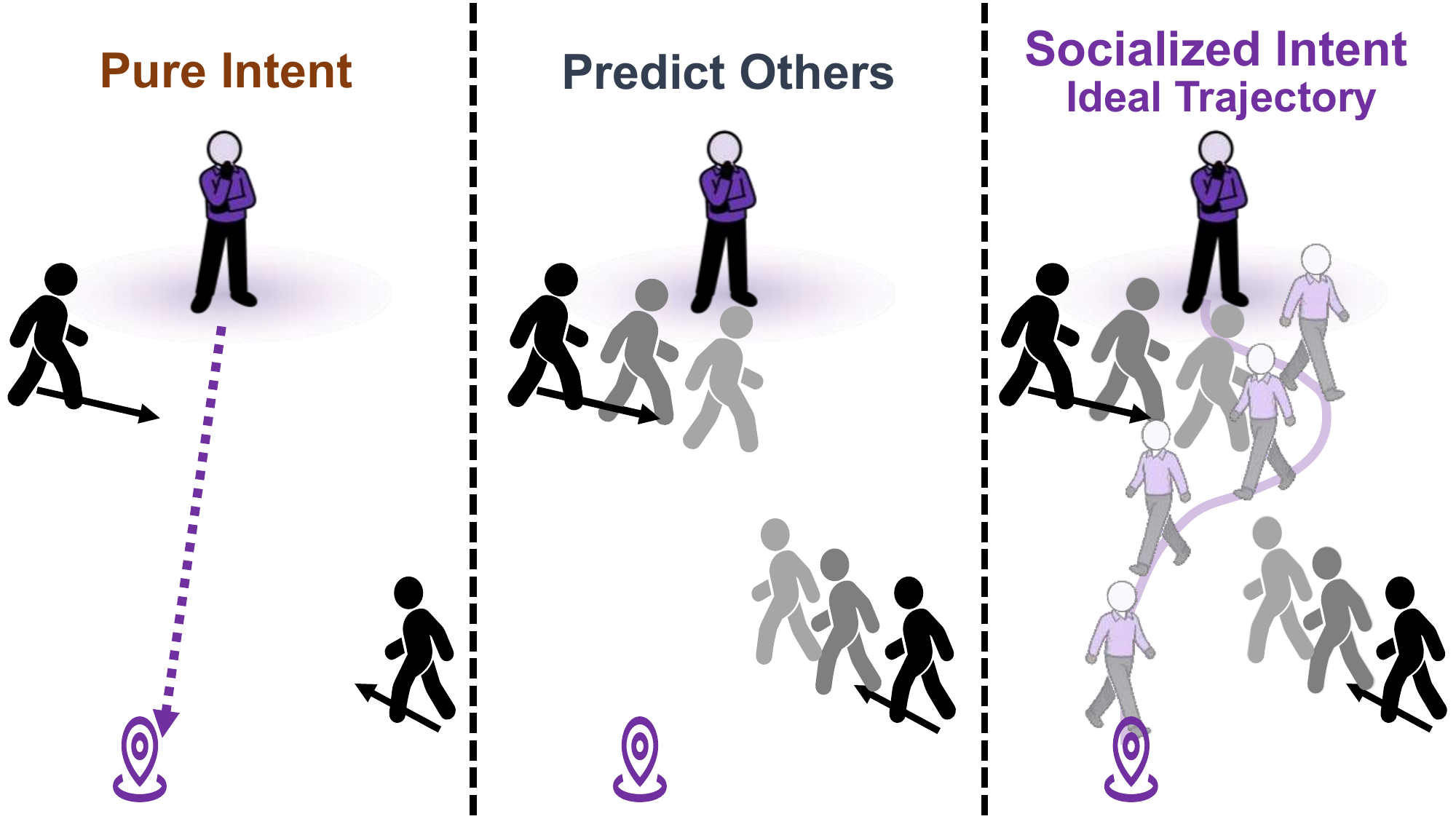}
  \caption{Illustration of socially shaped planning. Starting from pure intent, the agent anticipates others’ future motions and forms a socialized ideal trajectory before action.}
  \Description{Illustration of socially shaped planning. Starting from pure intent, the agent anticipates others’ future motions and forms a socialized ideal trajectory before action.}
  \label{fig:intro}
\end{figure}

Recent studies suggest that explicit decomposition provides a more suitable formulation for trajectory prediction. Prior work has supported this direction through high-level goal modeling, motion hierarchy decoupling, and structured representations in the time and frequency domains \cite{chen_2025_socialmoif,mangalam_2020_it,wong_2025_resonance,zhang_2024_demo}. These results show that separating the factors previously absorbed by a unified predictor is both reasonable and effective. However, most existing structured methods still treat social influence as part of a unified prediction process. They do not distinguish its functional role in trajectory formation. More specifically, the high-level representations used in existing decomposition or intent modeling usually remain at the level of target endpoints \cite{mangalam_2020_it}, motion modes \cite{zhang_2024_demo}, or abstract latent variables \cite{zhu_2023_biff}. They rarely correspond explicitly to a socially shaped planning representation formed after anticipating the future behaviors of surrounding agents.

In real scenarios, an agent often forms a socially constrained overall plan before executing a motion decision. This plan already reflects its anticipation of the future behaviors of surrounding agents, as shown in Fig.~\ref{fig:intro}. During execution, the agent further adjusts this plan in response to local interaction changes. The former determines the overall path trend and passing timing. It belongs to the planning stage. The latter appears as local trajectory deviations and immediate adjustments. It belongs to the reaction stage. This distinction suggests that trajectory formation should be modeled as an intent-driven two-stage process, in which socially shaped planning and local reaction play distinct roles. Existing decoupling methods have provided useful insight for structured prediction, but they still do not offer a unified framework for representing the agent's complete plan.

To capture this intent-driven process, we propose INTraJ, a unified social influence modeling framework. We term it INTraJ to highlight intent-driven trajectory prediction. It explicitly decomposes social influence in trajectory prediction into a planning stage and a reaction stage. We first extract pure intent from the target agent's history and local context as a self-intent planning seed. We then inject future social information derived from surrounding agents to obtain socialized intent, which yields a planning-level reference trajectory. It captures the target agent's overall passing plan after accounting for the future behaviors of surrounding agents. We then derive reaction-stage local adjustments from the difference between the full-context prediction and this planning reference, with a gating mechanism controlling their strength across samples and time steps. This same planning-and-reaction decomposition defines the core of INTraJ, and is instantiated under both multi-target \cite{zhou_2023_querycentric} and single-target prediction \cite{zhang_2024_demo} \cite{zhou_2022_hivt} paradigms with tailored designs within a unified formulation.

Autonomous driving and pure crowd scenarios are usually studied under different settings. Both still involve overall passing plans and local reactive adjustments in multi-agent interaction. Our method yields consistent gains on both tasks. This result indicates that hierarchical social interaction modeling generalizes across scenarios.

The main contributions are as follows:
\begin{enumerate}
\item We introduce a hierarchical social influence modeling view that decomposes social influence into a planning stage and a reaction stage, formulating trajectory formation as socially shaped planning followed by local reactive adjustment.
\item We instantiate the INTraJ framework under both the multi-target and single-target prediction paradigms with tailored designs within a unified formulation.
\item INTraJ delivers consistent gains across multiple backbones and benchmarks in both autonomous driving and crowd forecasting, achieving state-of-the-art performance in several settings. These results further show that explicitly decomposing social influence into planning and reaction provides an effective and generalizable paradigm for trajectory prediction.
\end{enumerate}

\section{Related Work}


Crowd trajectory prediction is a core setting for social interaction modeling. Early methods modeled neighbor influence and short-term interactions. Social LSTM \cite{alahi_2016_social} used explicit social pooling, and Social-STGCNN \cite{mohamed_2020_socialstgcnn} extended this line with spatiotemporal graph convolution. Later work further moved toward stronger unified modeling and richer interaction representations, such as agent-aware attention \cite{yuan_2021_agentformer}, urban crowd traffic modeling \cite{uhlemann_2025_snapshot}, generative formulations \cite{fu_2025_moflow}, time-frequency representations \cite{wong_2025_resonance}, and higher-order interaction intent \cite{chen_2025_socialmoif}. These studies move the field from local interaction modeling toward global representations, hierarchical modeling, and structured interaction modeling. 
Recent work such as SoPerModel~\cite{yang_sopermodel_2025} further explores social perception by learning social relationships among agents and incorporating the extracted social influence into trajectory prediction. However, these methods focus on representing social interaction patterns, whereas INTraJ investigates how social influence contributes to trajectory formation through a planning-before-reaction decomposition.


Trajectory prediction in autonomous driving involves HD maps, lane topology, and complex relations among traffic participants, which has driven the field from local interaction reasoning toward stronger scene-level joint representations. VectorNet \cite{gao_2020_vectornet} showed the effectiveness of vectorized map and agent encoding, while later scene-centric and query-based methods further strengthened unified scene modeling. QCNet \cite{zhou_2023_querycentric} and MTR \cite{shi_2022_motion} improved motion forecasting through query-centric scene modeling and the combination of global intention modeling with local refinement. More recent work has moved toward future-aware interaction and structured high-level modeling, for example by injecting potential future trajectories into scene encoding \cite{li_2025_futureaware}, jointly modeling high-level future intentions and low-level future behaviors \cite{zhu_2023_biff}, or decoupling motion forecasting into directional intentions and dynamic states \cite{zhang_2024_demo}. Overall, this line pushes autonomous driving prediction beyond direct trajectory regression toward richer scene reasoning and structured future modeling. Most methods still incorporate social interaction as part of unified scene reasoning or high-level conditioning, rather than modeling its distinct roles in planning and reaction.

\begin{figure*}[t] 
  \centering
  \includegraphics[width=480pt]{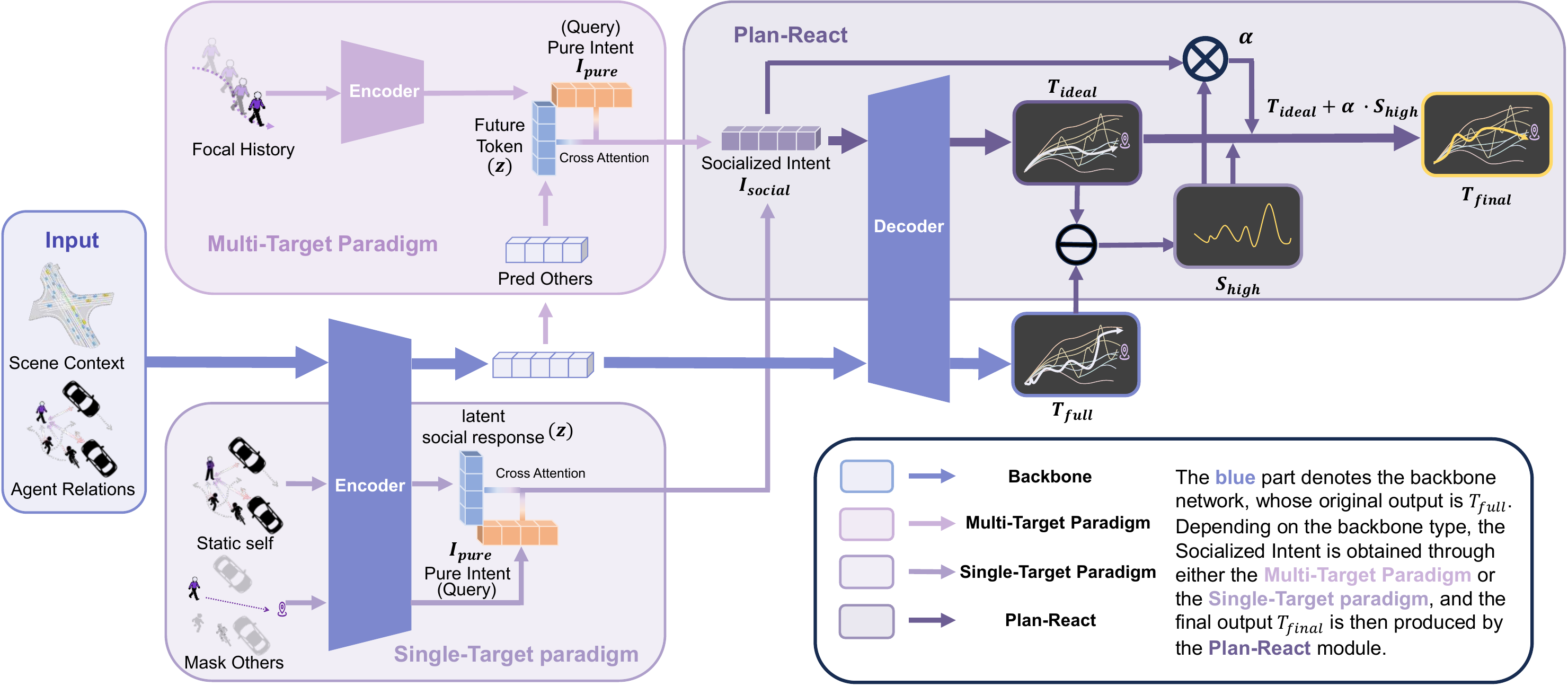} 
  \caption{Overview of our INTraJ framework. INTraJ unifies the Multi-Target and Single-Target paradigms under a two-stage formulation: it first constructs a socially shaped planning reference from the agent’s base intent and future social cues, and then injects local reactive corrections to produce the final trajectory. This decomposition separates global passing tendency from short-term interaction response.}
  \Description{Overview of our INTraJ framework. INTraJ unifies the Multi-Target and Single-Target paradigms under a two-stage formulation: it first constructs a socially shaped planning reference from the agent’s base intent and future social cues, and then injects local reactive corrections to produce the final trajectory. This decomposition separates global passing tendency from short-term interaction response.}
  \label{fig:work}
\end{figure*}


Recent work more directly related to our method spans structured high-level modeling, counterfactual or causal formulations, and plug-in refinement. One direction represents high-level intent as target endpoints, motion modes, or latent variables; PECNet \cite{mangalam_2020_it}, for example, uses a distant endpoint as a high-level condition for multimodal future generation, while later work further explored dual-level representations \cite{zhang_2024_demo}, higher-order intention modeling \cite{chen_2025_socialmoif}, and intention-guided reasoning \cite{pei_2025_foresight}. Another line models interaction effects through counterfactual \cite{alahi_2016_social} or causal formulations \cite{pourkeshavarz_2024_cadet}, mainly focusing on interaction attribution, stability, and disentanglement. Plug-in refinement methods instead emphasize compatibility with existing backbones through refinement or adaptation schemes, as exemplified by SmartRefine \cite{zhou_2024_smartrefine} and LaKD \cite{li_2024_lakd}. In contrast, our method starts from the trajectory formation process and explicitly models socially shaped planning followed by reaction-stage adjustment.

\section{Method}

In this section, we present the proposed hierarchical social influence modeling framework. Specifically, this approach decouples future trajectory formation into a planning stage and a reaction stage, providing tailored designs for both multi-target and single-target prediction paradigms. We first formulate the prediction task, and then detail the unified framework along with its specific implementations for each paradigm.

\subsection{Problem Formulation}

Given the historical trajectory of the agent $\mathbf{X}_{ego}$, the set of neighboring trajectories $\mathbf{X}_{nbr}$, and the scene context $\mathbf{C}$, trajectory prediction aims to model its future trajectory distribution $\mathbf{Y}$. For prediction scenarios, the model typically outputs $K$ candidate future trajectories along with their corresponding confidence scores, with each trajectory spanning $T_f$ future time steps.

The scene context $\mathbf{C}$ can include map elements, lane geometries, scene topology, and state information of other traffic participants. Despite differences in input organization and representation formats, the multi-target prediction paradigm typically models the map and multiple agents jointly within a unified scene representation, enabling the simultaneous output of future trajectories for multiple agents. In contrast, the single-target prediction paradigm centers its output on the future predictions of the focal agent (or selected agents), even though its conditioning information can still fuse the full scene context. Their shared core objective is to provide multi-modal future trajectory predictions for the focal agent, conditioned on its historical motion and the scene context.

Within this unified task formulation, this paper focuses on the structured modeling of social influence.

\subsection{Plan and Reaction Decomposition}

This subsection defines the unified Plan-React module of INTraJ. As shown in Fig.~\ref{fig:work}, given a pure planning seed $I_{pure}$ and a future social carrier $z$, the module first updates the planning representation to obtain $I_{social}$ and decodes the planning-level reference $T_{ideal}$, and then recovers local reactive adjustments to produce the final trajectory $T_{final}$.

First, we extract the pure intent representation $I_{pure}$ from the backbone encodings to serve as the planning seed. We then integrate the future social carrier $z$ to update this seed, yielding the socialized planning representation:
\begin{equation}
    I_{social} = \mathcal{F}(I_{pure}, z)
    \label{eq:isocial}
\end{equation}
where $\mathcal{F}(\cdot)$ denotes the planning-level social shaping module. The decoder generates the planning-level reference trajectory from $I_{social}$:
\begin{equation}
    T_{{ideal}} = \mathrm{Decoder}(I_{social}).
\end{equation}
$T_{\text{ideal}}$ captures the smooth, low-frequency overall motion plan under explicit future social constraints.

The reaction stage recovers local reactive adjustments, such as immediate collision avoidance and short-term speed changes. We retain the backbone's prediction capability under the full input context to generate the full prediction:
\begin{equation}
    T_{{full}} = \mathrm{Decoder}(\mathbf{X}_{ego}, \mathbf{X}_{nbr}, \mathbf{C}).
\end{equation}
We define the structural difference between the two predictions as:
\begin{equation}
    S_{{high}} = T_{{full}} - T_{{ideal}}.
\end{equation}
$S_{\text{high}}$ approximates the local reactive adjustments that deviate from the overall motion plan.

The requirement for local adjustments varies across samples, modes, and future time steps. We introduce a gating mechanism to selectively fuse the reactive adjustments:
\begin{equation}
    T_{{final}} = T_{{ideal}} + \alpha \cdot S_{{high}}.
\end{equation}
The learnable gating weight $\alpha$ controls the scale of reactive adjustments retained for each mode and time step.

Both multi-target and single-target prediction paradigms share this same Plan-React module. Their difference lies only in how the pure intent $I_{pure}$ and the future social carrier $z$ are constructed before entering the shared planning-to-reaction process. Therefore, the next two subsections focus on how the two paradigms instantiate $I_{pure}$ and $z$, while the subsequent ideal decoding, residual extraction, and gated fusion remain common.


\subsection{Multi-target Prediction Paradigm}

In the multi-target paradigm, the shared Plan-React module constructs the pure intent $I_{pure}$ from the focal agent and the future social carrier from the predicted futures of non-focal agents. Since multi-target backbones explicitly predict surrounding agents, their future trajectories naturally instantiate the social carrier.

We construct a lightweight self-history intent branch exclusively for the focal agent, extracting a planning seed from its own historical trajectory, denoted as pure intent $I_{pure}$. This seed retains the basic planning tendency derived from the focal agent's history and local context, without explicitly attending to the future behaviors of non-focal agents, and therefore serves as the reference starting point for the planning-stage update. Using the full-scene focal token directly as the planning reference would introduce a mixed encoding of map, neighbor interactions, and scene-level context, making it difficult to maintain the separation between pure intent and socially shaped intent.

Given $I_{pure}$, we first predict the future trajectories of non-focal agents from their scene-level representations, denoted as $P$, and then encode them into the future social carrier:

\begin{equation}
    z_{{future}} = \text{Encoder}(P_{{others}}).
\end{equation}
Here, $z_{{future}}$ instantiates the generic social carrier $z$ in Section~3.2. It explicitly carries prospective motion information of surrounding agents. We then use $I_{pure}$ as the query and $z_{{future}}$ as the key and value to aggregate future information through cross-attention:
\begin{equation}
    \Delta_{{social}} = \text{CrossAttn}(q = I_{pure},\ k=v=z_{{future}}),
\end{equation}
\begin{equation}
    I_{social} = I_{pure} + \Delta_{{social}}.
\end{equation}
$I_{social}$ is the updated form of the same planning seed after explicitly incorporating the future behaviors of others. It is the output of the planning-stage update and serves as the ideal-context representation for generating $T_{ideal}$.

Following the shared Plan-React module in Section~3.2, the ideal decoder then maps $I_{social}$ to the planning-level reference trajectory:
\begin{equation}
    T_{{ideal}} = \text{Decoder}_{\text{ideal}}(I_{social},\mathbf{C}),
\end{equation}
where the decoder condition retains the scene context provided by the multi-target backbone. $T_{ideal}$ is the reference path formed by the focal agent after explicitly accounting for the future behaviors of others at the planning level, emphasizing overall passage tendency, long-term navigability, and mode-level planning consistency.

Once $I_{pure}$ and the multi-target social carrier have been constructed in this way, the subsequent full prediction, residual extraction, and gated fusion all follow the unified Plan-React definitions in Section~3.2.

\subsection{Single-target Prediction Paradigm}
The single-target paradigm instantiates the shared Plan-React module differently. Instead of relying on explicit future prediction of surrounding agents, it constructs $I_{pure}$ from an ego-centered reference branch and instantiates the social carrier with a latent social response. Under this setting, we instantiate INTraJ on DeMo \cite{zhang_2024_demo}, whose decoupled design naturally supports separate construction of these two ingredients before the shared planning-stage update.

Unlike the multi-target paradigm, the single-target setting does not directly expose predicted futures of non-focal agents. We therefore construct the two inputs of the Plan-React module through auxiliary branches built on top of the original backbone.

The \textbf{mask others} branch keeps only the ego agent and map inputs, extracting the target agent’s base planning representation $I_{pure}$ without explicit neighbor conditions. This token encodes the ego history, local map constraints, and base motion tendency.

The \textbf{static self} branch sets the ego agent's historical motion to static while retaining its original position and type in the scene. This removes the ego agent's active motion increment and allows the model to encode potential future responses of surrounding agents under a static-ego condition. The resulting latent preserves scene relations and instantiates the generic social carrier $z_{{latent}}$.

We then use $I_{pure}$ as the query and $z_{{latent}}$ as the key and value. Cross-attention injects this latent social response into the planning representation:
\begin{equation}
I_{social} = \mathrm{CrossAttn}(q = I_{pure},\ k=v=z_{{latent}}).
\end{equation}

This operation injects the latent social response into the reference plan. In this way, the single-target paradigm replaces explicit future prediction with a latent social response while preserving the same planning-stage update.

Following the shared Plan-React module in Section~3.2, the ideal branch then maps $I_{social}$ to the planning-level reference trajectory:
\begin{equation}
    T_{{ideal}} = \text{Decoder}_{\text{ideal}}(I_{social}).
\end{equation}
This branch is kept at the same decoding level as the original single-target backbone so that $T_{ideal}$ preserves the full semantics of the planning-level reference. After incorporating the latent social response, $T_{ideal}$ captures the overall passing trend of the target agent.

\begin{table*}[t]
  \caption{Comparisons to other state-of-the-art methods on ETH-UCY \cite{alahi_2016_social} (left) and SDD \cite{andle_2023_stanford} (right). Metrics are ADE/FDE (best-of-20), in meters on ETH-UCY and in pixels on SDD. Lower metrics indicate better performance. \highlight{Purple} numbers mark the top 2 results on each set.}
  \label{tab:crowd_benchmarks}
  \centering
  \footnotesize
  \setlength{\tabcolsep}{4pt}
  \renewcommand{\arraystretch}{1.08}
  \begin{minipage}[t]{0.72\textwidth}
    \begin{tabular*}{\linewidth}{@{\extracolsep{\fill}}lcccccc@{}}
      \toprule
      Method (ETH-UCY) & eth$\downarrow$ & hotel$\downarrow$ & univ$\downarrow$ & zara1$\downarrow$ & zara2$\downarrow$ & Average$\downarrow$ \\
      \midrule
      MS-TIP \cite{chib_2024_mstip} (2024) & 0.39 / 0.57 & 0.13 / 0.22 & 0.24 / 0.40 & 0.20 / 0.34 & 0.17 / 0.29 & 0.22 / 0.36 \\
      SMEMO \cite{marchetti_2024_smemo} (2024) & 0.39 / 0.59 & 0.14 / 0.20 & 0.23 / 0.41 & 0.19 / 0.32 & 0.15 / 0.26 & 0.22 / 0.35 \\
      Trajectron++ (2020) \cite{salzmann_2020_trajectron} & 0.43 / 0.86 & 0.12 / 0.19 & 0.22 / 0.43 & 0.17 / 0.32 & \highlight{\textbf{0.12}} / 0.25 & 0.20 / 0.39 \\
      LG-Traj (2024) \cite{chib_2025_lgtraj} & 0.38 / 0.56 & \highlight{\textbf{0.11}} / 0.17 & 0.23 / 0.42 & 0.18 / 0.33 & 0.14 / 0.25 & 0.20 / 0.34 \\
      PPT (2024) \cite{lin_2024_progressive} & 0.36 / 0.51 & \highlight{\textbf{0.11}} / \highlight{\textbf{0.14}} & 0.22 / 0.40 & 0.17 / 0.30 & \highlight{\textbf{0.12}} / \highlight{\textbf{0.21}} & 0.20 / 0.31 \\
      E-V$^2$-Net (2025) \cite{xia_2025_another} & 0.25 / 0.38 & \highlight{\textbf{0.11}} / 0.16 & 0.23 / 0.42 & 0.19 / 0.30 & 0.13 / 0.24 & \highlight{\textbf{0.18}} / 0.30 \\
      AgentFormer (2021) \cite{yuan_2021_agentformer} & 0.26 / 0.39 & \highlight{\textbf{0.11}} / \highlight{\textbf{0.14}} & 0.26 / 0.46 & \highlight{\textbf{0.15}} / \highlight{\textbf{0.23}} & 0.14 / 0.23 & \highlight{\textbf{0.18}} / 0.29 \\
      SocialCircle (2024) \cite{wong_2024_socialcircle} & 0.25 / 0.38 & 0.12 / \highlight{\textbf{0.14}} & 0.23 / 0.42 & 0.18 / 0.29 & 0.13 / 0.22 & \highlight{\textbf{0.18}} / 0.29 \\
      Y-net (2021) \cite{mangalam_2021_goals} & 0.28 / \highlight{\textbf{0.33}} & \highlight{\textbf{0.10}} / \highlight{\textbf{0.14}} & 0.24 / 0.41 & 0.17 / 0.27 & 0.13 / 0.22 & \highlight{\textbf{0.18}} / \highlight{\textbf{0.27}} \\
      UPDD (2024) \cite{liu_2024_uncertaintyaware} & \highlight{\textbf{0.22}} / 0.42 & 0.17 / 0.30 & \highlight{\textbf{0.14}} / \highlight{\textbf{0.28}} & \highlight{\textbf{0.16}} / 0.30 & 0.14 / 0.31 & \highlight{\textbf{0.17}} / 0.32 \\
      TAMLD (2025) \cite{ren_2025_totp} & 0.39 / 0.58 & 0.13 / 0.18 & 0.22 / 0.37 & 0.18 / 0.28 & 0.13 / \highlight{\textbf{0.21}} & 0.21 / 0.32 \\
      IAD (2026) \cite{liu_2026_intentionaware} & 0.34 / 0.52 & 0.15 / 0.24 & 0.20 / 0.36 & \highlight{\textbf{0.15}} / \highlight{\textbf{0.24}} & \highlight{\textbf{0.11}}  / \highlight{\textbf{0.20}} & 0.19 / 0.31 \\
      Resonance (2025) \cite{wong_2025_resonance} & \highlight{\textbf{0.23}} / \highlight{\textbf{0.35}} & \highlight{\textbf{0.10}} / \highlight{\textbf{0.15}} & 0.24 / 0.41 & 0.17 / 0.29 & 0.13 / 0.22 & \highlight{\textbf{0.17}} / 0.28 \\
      \midrule
      \textit{Ours (Resonance)} & \highlight{\textbf{0.23}} / \highlight{\textbf{0.35}} & \highlight{\textbf{0.10}} / \highlight{\textbf{0.15}} & 0.23 / \highlight{\textbf{0.34}} & 0.17 / 0.29 & 0.14 / \highlight{\textbf{0.21}} & \highlight{\textbf{0.17}} / \highlight{\textbf{0.26}} \\
      \bottomrule
    \end{tabular*}
  \end{minipage}%
  \hspace{0.008\textwidth}%
  \begin{minipage}[t]{0.24\textwidth}
    \begin{tabular*}{\linewidth}{@{\extracolsep{\fill}}lc@{}}
      \toprule
      Method (SDD) & ADE/FDE$\downarrow$ \\
      \midrule
      FlowChain (2023) \cite{maeda_2023_fast} & 9.93 / 17.17 \\
      IMP (2023) \cite{shi_2023_representing} & 8.98 / 15.54 \\
      SMEMO (2024) \cite{marchetti_2024_smemo} & 8.11 / 13.06 \\
      LG-Traj (2024) \cite{chib_2025_lgtraj} & 7.80 / 12.79 \\
      Y-net (2021) \cite{mangalam_2021_goals} & 7.85 / 11.85 \\
      UEN (2024) \cite{su_2024_unified} & 7.30 / 10.40 \\
      PPT (2024) \cite{lin_2024_progressive} & 7.03 / 10.65 \\
      UPDD (2024) \cite{liu_2024_uncertaintyaware} & 6.59 / 13.90 \\
      E-V$^2$-Net (2025) \cite{xia_2025_another} & 6.57 / 10.49 \\
      SocialCircle (2024) \cite{wong_2024_socialcircle} & 6.54 / 10.36 \\
      MUSE-VAE (2022) \cite{lee_2022_musevae} & 6.36 / 11.10 \\
      IAD (2026) \cite{liu_2026_intentionaware} & 6.85 / 11.22 \\
      Resonance (2025) \cite{wong_2025_resonance} & 6.27 / 10.02 \\
      \midrule
      \textit{Ours (Resonance)} & \highlight{\textbf{6.19}} / \highlight{\textbf{9.87}} \\
      \bottomrule
    \end{tabular*}
  \end{minipage}
\end{table*}

Compared with the multi-target design, the distinctive part here lies in how the social carrier is obtained: it is derived from a counterfactual latent response rather than explicit future prediction. After this paradigm-specific construction, the resulting $I_{social}$ enters the same Plan-React pipeline, so the final prediction is still produced through the common ideal-reference and reaction-fusion process.

\subsection{Training Objectives}

During training, we jointly optimize the final prediction branch and the planning reference branch. The total loss consists of the final prediction loss, the planning reference loss, and an auxiliary regularization term.

The prediction loss for the final output trajectory $T_{final}$, denoted $\mathcal{L}_{\text{pred}}$, remains the same as in the original backbone. For multimodal prediction, $\mathcal{L}_{\text{pred}}$ follows the backbone's original best-of-$K$ regression loss, mode classification loss, or their combination. This preserves the original prediction formulation.

For the planning-level reference trajectory $T_{ideal}$, we introduce a planning reference loss $\mathcal{L}_{\text{ref}}$. We obtain a smoothed target trajectory $T_{smooth}$ by applying mild smoothing to the ground-truth future trajectory, and use it as the supervision target:
\begin{equation}
\mathcal{L}_{\text{ref}} = \mathcal{L}_{\text{traj}}(T_{ideal},\, T_{smooth}) + \lambda_{\text{end}}\,\mathcal{L}_{\text{end}}(T_{ideal},T_{smooth}),
\end{equation}
where $\mathcal{L}_{\text{traj}}$ is the trajectory regression loss, and $\mathcal{L}_{\text{end}}$ is the endpoint loss. The latter constrains the endpoint position and overall passing direction of the reference trajectory.

We further introduce an auxiliary loss $\mathcal{L}_{\text{aux}}$ to regulate the learning behavior of the gate. The final trajectory is defined as
\begin{equation}
T_{final} = T_{ideal} + \alpha \cdot S_{high}
\end{equation}
Without additional constraints, the model may drive $\alpha$ toward 1 in early training and collapse to the full prediction branch. We therefore impose a variance-based regularizer on the gate. This regularizer prevents early collapse to a uniformly high value and encourages dynamic responses across samples, modes, and time steps.

The overall loss is
\begin{equation}
\mathcal{L} = \mathcal{L}_{\text{pred}} + \lambda_1 \mathcal{L}_{\text{ref}} + \lambda_2 \mathcal{L}_{\text{aux}},
\end{equation}
where $\lambda_1$ and $\lambda_2$ are balancing coefficients. This objective remains compatible with the original backbone prediction target. It supports plug-in integration with different multi-target or single-target prediction models without changing the original prediction paradigm.

\section{Experiments}

\begin{figure*}[t] 
  \centering
  \includegraphics[width=460pt]{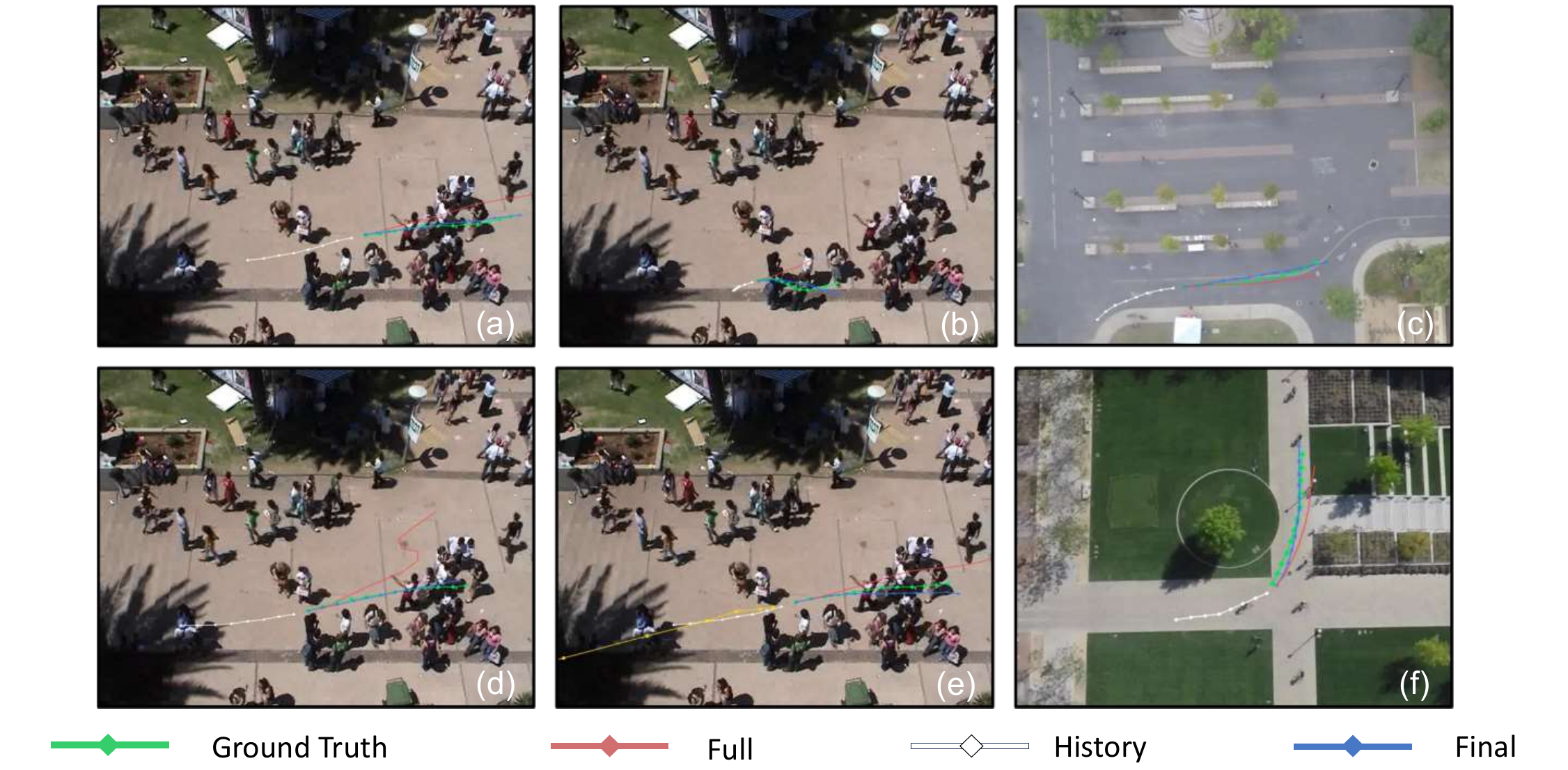} 
  \caption{Qualitative results on ETH/UCY. Compared with full, final alleviates the entanglement caused by coupled prediction (Subfigure d), avoiding being misled by spuriously correlated neighbor patterns while preserving a more coherent motion trend.}
  \Description{Qualitative results on ETH/UCY. Compared with full, final alleviates the entanglement caused by coupled prediction (Subfigure d), avoiding being misled by spuriously correlated neighbor patterns while preserving a more coherent motion trend.}
  \label{fig:crowd}
\end{figure*}

\begin{table*}[t]
  \caption{Comparison to state-of-the-art methods on the Argoverse 2 test set. Numbers with underlines mark the best results for each metric.}
  \label{tab:av2_main}
  \centering
  \footnotesize
  \begin{tabular*}{\linewidth}{@{\extracolsep{\fill}}lccccccc@{}}
    \toprule
    Method & \textbf{b-minFDE}$_6\downarrow$ & minFDE$_6\downarrow$ & minADE$_6\downarrow$ & MR$_6\downarrow$ & minFDE$_1\downarrow$ & minADE$_1\downarrow$ & MR$_1\downarrow$ \\
    \midrule
    THOMAS (2022) \cite{gilles_2022_thomas}  & 2.16 & 1.51 & 0.88 & 0.20 & 4.71 & 1.95 & 0.64 \\
    Forecast-MAE (2023) \cite{cheng_2023_forecastmae}  & 2.03 & 1.39 & 0.71 & 0.17 & 4.35 & 1.74 & 0.61 \\
    GoReIa (2023) \cite{cui_2023_gorela}  & 2.01 & 1.48 & 0.76 & 0.22 & 4.62 & 1.82 & 0.66 \\
    HeteroGCN (2023) \cite{gao_2023_dynamic} & 2.00 & 1.37 & 0.73 & 0.18 & 4.53 & 1.79 & 0.59 \\
    MTR (2022) \cite{shi_2022_motion} & 1.98 & 1.44 & 0.73 & 0.15 & 4.39 & 1.74 & 0.58 \\
    GANet (2023) \cite{wang_2023_ganet} & 1.96 & 1.34 & 0.72 & 0.17 & 4.48 & 1.77 & 0.59 \\
    DySeT (2024) \cite{pourkeshavarz_2024_dyset} & 1.93 & 1.28 & 0.67 & 0.16 & 4.41 & 1.76 & 0.61 \\
    FINet (2025) \cite{li_2025_futureaware} & 1.93 & 1.27 & 0.66 &0.15 & 4.02 & 1.60 & 0.57 \\
    \textcolor{qcnetpurple}{QCNet (2023) \cite{zhou_2023_querycentric}}  & \textcolor{qcnetpurple}{1.91} & \textcolor{qcnetpurple}{1.29} & \textcolor{qcnetpurple}{0.65} & \textcolor{qcnetpurple}{0.16} & \textcolor{qcnetpurple}{4.30} & \textcolor{qcnetpurple}{1.69} & \textcolor{qcnetpurple}{0.59} \\
    ProphNet (2023) \cite{wang_2023_prophnet} & 1.88 & 1.32 & 0.66 & 0.18 & 4.77 & 1.76 & 0.61 \\
    SmartRefine (2024) \cite{zhou_2024_smartrefine}  & 1.86 & 1.23 & 0.63 & 0.15 & 4.17 & 1.65 & 0.58 \\
    \textcolor{demopurple}{DeMo (2024) \cite{zhang_2024_demo}}  &  \textcolor{demopurple}{1.84} &  \textcolor{demopurple}{1.17} &  \textcolor{demopurple}{\textbf{\underline{0.61}}} &  \textcolor{demopurple}{\textbf{\underline{0.13}}} &  \textcolor{demopurple}{3.74} &  \textcolor{demopurple}{1.49} &  \textcolor{demopurple}{0.55} \\
    DONUT (2025) \cite{knoche_2025_donut} & \textbf{\underline{1.79}} & \textbf{\underline{1.16}} & 0.63 & 0.14 & 4.06 & 1.66 & \textbf{\underline{0.54}} \\
    \midrule
    \textcolor{qcnetpurple}{\textbf{Ours (QCNet)}} & \textcolor{qcnetpurple}{1.87} & \textcolor{qcnetpurple}{1.25} & \textcolor{qcnetpurple}{0.64} & \textcolor{qcnetpurple}{0.15} & \textcolor{qcnetpurple}{4.14} & \textcolor{qcnetpurple}{1.63} & \textcolor{qcnetpurple}{0.56} \\
     \textcolor{demopurple}{\textbf{Ours (DeMo)}} &  \textcolor{demopurple}{1.83} &  \textcolor{demopurple}{1.17} &  \textcolor{demopurple}{\textbf{\underline{0.61}}} &  \textcolor{demopurple}{\textbf{\underline{0.13}}} &  \textcolor{demopurple}{\textbf{\underline{3.71}}} &  \textcolor{demopurple}{\textbf{\underline{1.48}}} &  \textcolor{demopurple}{0.55} \\
    \bottomrule
  \end{tabular*}
\end{table*}

\subsection{Experimental Setup}

We evaluate INTraJ on four benchmarks: Argoverse 2 (AV2) \cite{wilson_2021_argoverse} and Argoverse 2-ped \cite{uhlemann_2025_snapshot} for autonomous driving, and ETH/UCY \cite{alahi_2016_social} and the Stanford Drone Dataset \cite{andle_2023_stanford} (SDD) for pure crowd trajectory prediction. For evaluation, we report ADE and FDE on ETH/UCY, SDD, and AV2-ped. On AV2, we follow the standard evaluation protocol and report ADE/FDE together with the official metrics MR6 and MR1. The official leaderboard is no longer open for submission, so we cannot report online results. All AV2 experiments in this paper follow the current public evaluation settings used in prior work.

To evaluate the generality of INTraJ across different modeling paradigms and backbones, we implement it on four backbones. On AV2, we use QCNet \cite{zhou_2023_querycentric} and DeMo \cite{zhang_2024_demo} as representatives of the multi-target and single-target prediction paradigms, respectively. On ETH/UCY \cite{alahi_2016_social} and SDD \cite{andle_2023_stanford}, we use Resonance. On AV2-ped, we further build a lightweight baseline with a small Transformer encoder and a small TCN decoder, and integrate INTraJ directly into this model. Unless otherwise noted, all experiments follow the original training and inference protocols of the corresponding backbones to ensure fair comparison.

\subsection{Results on Crowd Benchmarks}

We first evaluate INTraJ on pure crowd trajectory prediction benchmarks. The results are shown in Table \ref{tab:crowd_benchmarks}. The table summarizes a set of representative methods from 2024 and 2025 on ETH/UCY \cite{alahi_2016_social} and SDD \cite{andle_2023_stanford}, and further reports the performance of Resonance after integrating INTraJ. Overall, INTraJ delivers consistent gains on both crowd backbones. With Resonance as the backbone, it achieves state-of-the-art performance.

On SDD, the FDE of existing methods generally remains above 10. After integrating INTraJ, this metric drops to 9.87, showing a clear advantage. On ETH/UCY, INTraJ also delivers consistent gains. The overall FDE improves by 7.2\% over the corresponding backbone. On the ETH/UCY subset that emphasizes long-horizon prediction, the FDE gain reaches 12.2\%. These results show that the proposed planning-stage and reaction-stage decomposition effectively models complex interactions and long-horizon trajectory evolution in pure crowd scenarios.

Fig.~\ref{fig:crowd} shows qualitative results on ETH/UCY. Instead of displaying only the final prediction, we visualize two trajectories, $\texttt{full}$ and $\texttt{final}$. Here, $\texttt{full}$ denotes the base prediction under full input conditions, and $\texttt{final}$ denotes the final output. Compared with $\texttt{full}$, $\texttt{final}$ better resolves local interactions while preserving a coherent global motion trend. This yields predictions that better match real crowd dynamics. The same pattern appears in Table \ref{tab:crowd_benchmarks}, where INTraJ improves FDE by 12.2\% on the ETH/UCY long-horizon subset. These results show that INTraJ models crowd dynamics and long-horizon motion trends effectively.

\subsection{Results on Autonomous Driving Benchmarks}

We further evaluate INTraJ on autonomous driving trajectory prediction benchmarks. Table~\ref{tab:av2_main} compares INTraJ with a set of methods from 2024 and 2025 on AV2 \cite{wilson_2021_argoverse}, and reports the performance changes after integrating INTraJ into QCNet \cite{zhou_2023_querycentric} and DeMo \cite{zhang_2024_demo}. Overall, INTraJ achieves competitive results on AV2. It does not attain the best score on every metric, but it consistently improves both representative backbones. This result shows that the framework remains compatible with both the multi-target and single-target prediction paradigms. We choose QCNet and DeMo as representative backbones for two reasons. First, they correspond to the two paradigms studied in our method. Second, several 2025 methods are not publicly available, or their released implementations are incomplete, which makes stable reproduction of the reported results difficult. They are therefore not suitable for fair plug-in evaluation.

\begin{figure*}[t] 
  \centering
  \includegraphics[width=440pt]{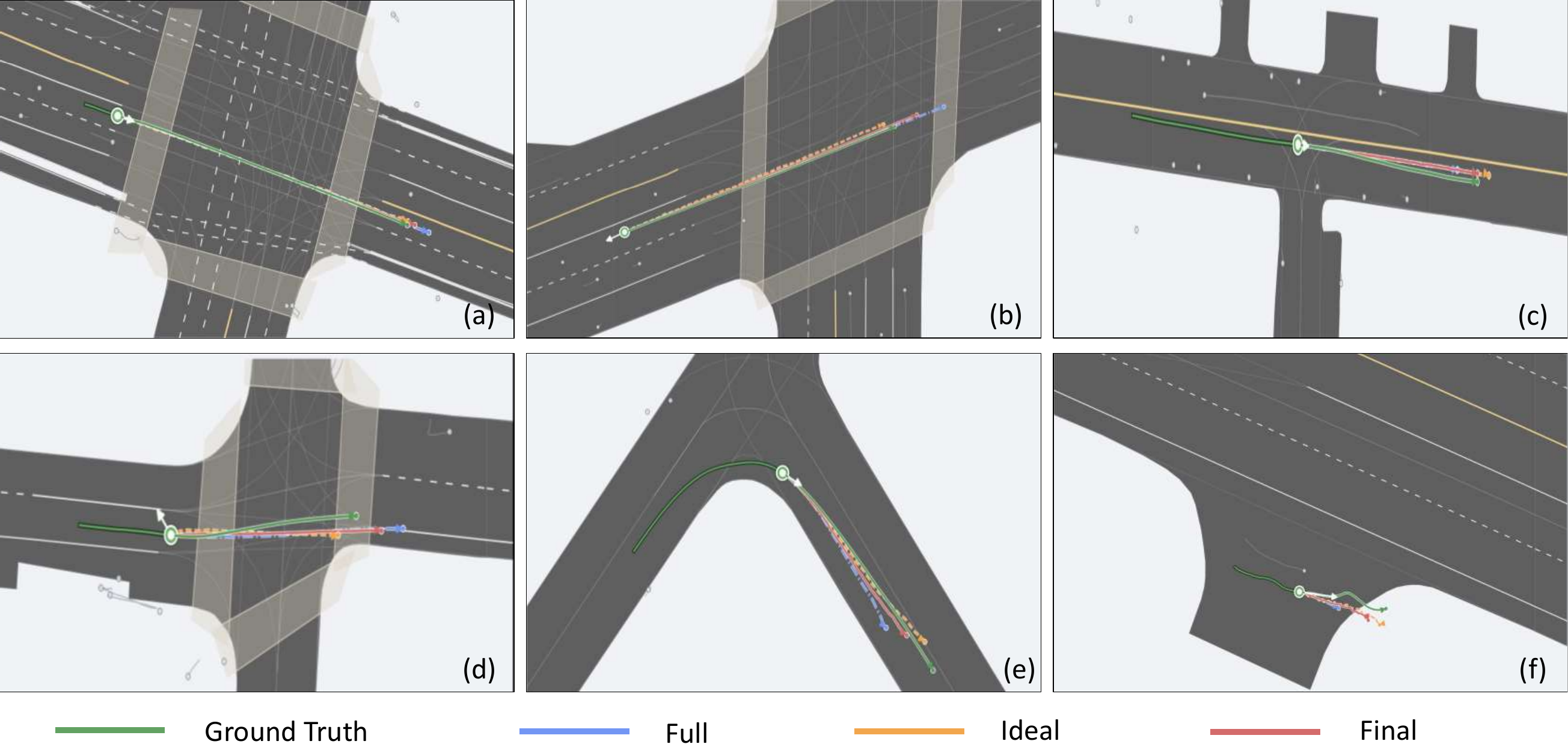} 
  \caption{Qualitative results on AV2. ideal is the planning-level reference. full denotes the backbone prediction with full input. final adds reactive corrections to the planning trend. This produces more realistic and adaptive trajectories.}
  \Description{Qualitative results on AV2. ideal is the planning-level reference. full denotes the backbone prediction with full input. final adds reactive corrections to the planning trend. This produces more realistic and adaptive trajectories.}
  \label{fig:auto}
\end{figure*}

\begin{table}[t]
  \caption{ADE and FDE values on the proposed benchmark test split, evaluating only the most likely predictions. Lower is better.}
  \label{tab:av2_ped}
  \centering
  
  \setlength{\tabcolsep}{5pt}      
  \renewcommand{\arraystretch}{1.2} 
  \small                             
  
  \begin{tabularx}{200pt}{@{}X@{\hspace{-50pt}}c@{\hspace{50pt}}c@{}}
    \toprule
    Models & ADE in m & FDE in m \\
    \midrule
    CVM \cite{uhlemann_2024_evaluating} & 0.793 & 1.776 \\
    SIMPL \cite{zhang_2024_simpl} & 0.699 & 1.557 \\
    Forecast-MAE \cite{cheng_2023_forecastmae} & 0.698 & 1.435 \\
    QCNet \cite{zhou_2023_querycentric} & 0.693 & 1.474 \\
    Snapshot \cite{uhlemann_2025_snapshot} & 0.567 & 1.251 \\
    Ours (Basic) & \textbf{0.558} & \textbf{1.228} \\
    \bottomrule
  \end{tabularx}
\end{table}

To verify that the method does not depend on a large backbone, we build a lightweight baseline on AV2-ped  \cite{uhlemann_2025_snapshot} and integrate INTraJ into it, as shown in Table~\ref{tab:av2_ped}. Although the baseline uses only a small Transformer encoder and a small TCN decoder, INTraJ still achieves state-of-the-art performance on all metrics. This result shows that the gains come from the structural decomposition into the planning stage and the reaction stage, rather than from coupling with a specific high-performance backbone.

Fig.~\ref{fig:auto} shows qualitative results on AV2. In this visualization, we present three trajectories: $\texttt{ideal}$, $\texttt{full}$, and $\texttt{final}$. The $\texttt{ideal}$ trajectory is the planning-level reference. It highlights the overall passing path and long-horizon traversability. The $\texttt{final}$ trajectory further incorporates reactive adjustments caused by local interaction changes. This progression from planning reference to final prediction illustrates how INTraJ decouples overall passing planning from local interaction response in autonomous driving scenarios.

\subsection{Comparison with Plug-in Refinement Methods}

\begin{table}[t]
  \caption{Comparison with plug-in refinement methods. Lower is better for minADE1, minFDE1, and MR1.}
  \label{tab:plugin_compare}
  \centering
  \scriptsize
  \setlength{\tabcolsep}{3.5pt}
  \renewcommand{\arraystretch}{1.2} 
  \small
  
  \begin{tabularx}{200pt}{@{}l@{\hspace{23pt}}c@{\hspace{23pt}}c@{\hspace{23pt}}c}
    \toprule
    Method & minADE1 & minFDE1 & MR1 \\
    \midrule
    QCNet \cite{zhou_2023_querycentric} & 1.69 & 4.30 & 0.59 \\
    + SmartRefine \cite{zhou_2024_smartrefine} & 1.65 & 4.17 & 0.58 \\
    + LAKD \cite{li_2024_lakd} & 1.66 & 4.15 & 0.58 \\
    + Ours & \textbf{1.63} & \textbf{4.14} & \textbf{0.56} \\
    \bottomrule
  \end{tabularx}
\end{table}

Beyond the main results, we directly compare INTraJ with the recently proposed plug-in post-processing method SmartRefine \cite{zhou_2024_smartrefine}. The results are shown in Table~\ref{tab:plugin_compare}. For a fair comparison, we evaluate these plug-in methods on the same QCNet backbone and compare the performance gains they provide.
The results show that INTraJ outperforms both SmartRefine~\cite{zhou_2024_smartrefine} and LAKD~\cite{li_2024_lakd} on QCNet~\cite{zhou_2023_querycentric}, yielding larger improvements across all three metrics.

More importantly, the advantage of INTraJ is not limited to numerical gains on a single benchmark. It also shows stronger cross-task generalization. Unlike plug-in methods that are mainly validated on the AV series, INTraJ delivers consistent gains in both autonomous driving and pure crowd scenarios. This result shows that its contribution is not tied to a specific dataset or task setting. Instead, it provides a more general structured formulation for trajectory formation across different interaction scenarios.

\subsection{Ablation Study}

\begin{table}[t]
  \caption{Ablation study of our method on the Argoverse 2 test set}
  \label{tab:ablation}
  \centering
  \scriptsize
  \setlength{\tabcolsep}{4pt}
  \renewcommand{\arraystretch}{1.2}
  \small
  
  \begin{tabularx}{200pt}{@{}l@{\hspace{21pt}}c@{\hspace{21pt}}c@{\hspace{21pt}}c@{\hspace{21pt}}c@{}}
    \toprule
    ideal & fusion & gate & minADE1 & minFDE1 \\
    \midrule
    \ding{55} & \ding{55} & \ding{55} & 1.69 & 4.30 \\
    \ding{51} & \ding{55} & \ding{55} & 1.75 & 4.29 \\
    \ding{51} & \ding{51} & \ding{55} & 1.69 & 4.30 \\
    \ding{51} & \ding{51} & \ding{51} & \textbf{1.63} & \textbf{4.14} \\
    \bottomrule
  \end{tabularx}
\end{table}

Finally, we conduct a systematic ablation study of the internal design with QCNet \cite{zhou_2023_querycentric} as the backbone. The results are shown in Table~\ref{tab:ablation}. We focus on four questions: whether the planning stage and reaction stage are both necessary, whether the carrier of future social influence is critical for constructing the planning-level reference, whether the gating mechanism can stably control reaction-stage adjustment injection, and how performance changes when the two-stage structure degrades to a unified prediction scheme.

The ablation results show that both the planning stage and the reaction stage are necessary. Removing the planning-level reference and keeping only the refinement branch prevents the model from stably capturing the overall passing trend. Removing the reaction-stage adjustment and relying only on the planning reference makes it difficult to absorb local interaction changes. Performance also drops when we remove the future social influence update or the gating mechanism. This result shows that the coordination among pure intent, the carrier of future social influence, and reactive residual injection is critical to the final performance. Overall, Table~\ref{tab:ablation} confirms the necessity of the two-stage design in INTraJ. It also shows that the gains do not come from additional parameter stacking, but from a more suitable formulation of trajectory formation.

\subsection{Computational Overhead Analysis}

Although INTraJ introduces explicit Plan-React decomposition and reaction correction modules, it is designed as a plug-in framework with moderate additional computational costs. We analyze the overhead in terms of parameter number, inference latency, and FLOPs.

As shown in Table~\ref{tab:efficiency}, compared with QCNet, INTraJ increases the parameters from 7.66M to 8.46M, latency from 37.94ms to 44.64ms, and FLOPs from 76.80G to 77.85G. The additional cost mainly comes from the future social carrier construction, planning-stage update, and gated reaction correction. Compared with SmartRefine, INTraJ achieves better prediction performance with comparable overhead, demonstrating a favorable accuracy-efficiency trade-off.

\begin{table}[t]
  \caption{Computational overhead comparison on the QCNet backbone}
  \label{tab:efficiency}
  \centering
  \scriptsize
  \setlength{\tabcolsep}{4pt}
  \renewcommand{\arraystretch}{1.2}
  \small
  
  \begin{tabularx}{200pt}{@{}l@{\hspace{18pt}}c@{\hspace{18pt}}c@{\hspace{18pt}}c@{}}
    \toprule
    Method & Params (M) & Lat. (ms) & FLOPs (G) \\
    \midrule
    QCNet & 7.66 & 37.94 & 76.80 \\
    + SmartRefine & 8.00 & 40.45 & 77.34 \\
    + INTraJ & 8.46 & 44.64 & 77.85 \\
    \bottomrule
  \end{tabularx}
\end{table}

\subsection{Robustness to Imperfect Pred Others}

In practical scenarios, future trajectories of surrounding agents inevitably contain prediction errors, while the future social carrier in INTraJ relies on these predicted trajectories. Therefore, we further evaluate the robustness of INTraJ under imperfect Pred Others. Specifically, we inject Gaussian noise with different standard deviations into Pred Others and measure the resulting performance changes.

\begin{figure}[t]
  \centering
  \includegraphics[width=\linewidth]{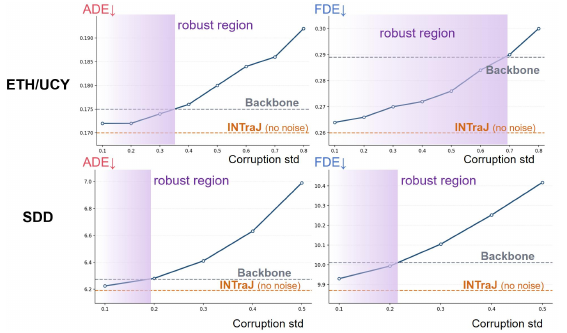}
  \caption{Robustness analysis under imperfect Pred Others. Gaussian noise with different standard deviations is injected into Pred Others. The shaded regions indicate corruption ranges where INTraJ remains better than or comparable to the backbone.}
  \Description{Robustness analysis under imperfect Pred Others. Gaussian noise with different standard deviations is injected into Pred Others. The shaded regions indicate corruption ranges where INTraJ remains better than or comparable to the backbone.}
  \label{fig:robustness}
\end{figure}

As shown in Fig.~\ref{fig:robustness}, INTraJ demonstrates strong tolerance to prediction corruption on both ETH/UCY and SDD. Under small perturbations, ADE and FDE only degrade slightly and remain consistently better than the backbone. As the corruption level increases, the performance decreases smoothly rather than collapsing abruptly, indicating that INTraJ can tolerate a reasonable level of uncertainty in social future prediction while maintaining the effectiveness of the Plan-React decomposition.

\section{Conclusion}

We presented INTraJ, a unified hierarchical social influence modeling framework that reformulates trajectory prediction as a two-stage process of planning and reaction. Instead of treating social interaction as a monolithic factor within a unified predictor, INTraJ distinguishes its functional roles in trajectory formation by constructing a planning-level reference and then modeling local reactive adjustments relative to it. This formulation remains applicable to both multi-target and single-target prediction paradigms and generalizes across autonomous driving and crowd forecasting scenarios. Experiments on multiple backbones and benchmarks show consistent improvements, with especially clear gains in final displacement error and long-horizon consistency, while achieving state-of-the-art results in several settings. These results highlight the value of staged social modeling for stable and coherent trajectory prediction.

\begin{acks}
This work was supported in part by the National Natural Science Foundation of China (Grant No. 62302086).
\end{acks}

\clearpage
\bibliographystyle{ACM-Reference-Format}

\bibliography{ref}

\end{document}